\pdfoutput=1

\documentclass[11pt]{article}

\usepackage[preprint]{acl}

\usepackage{times}
\usepackage{latexsym}

\usepackage[T1]{fontenc}

\usepackage[utf8]{inputenc}

\usepackage{microtype}

\usepackage{inconsolata}

\usepackage{graphicx}

\usepackage{amsmath}
\usepackage[capitalize,noabbrev]{cleveref}
\usepackage{graphicx,subfigure}
\usepackage{url}            
\usepackage{booktabs}       
\usepackage{nicefrac}       
\usepackage{microtype}      
\usepackage{xcolor}         
\usepackage{colortbl}
\usepackage{subcaption}
\usepackage{subfigure}
\usepackage{enumitem}
\usepackage{multirow}

\usepackage{makecell} 
\usepackage{amsfonts,amssymb}
\usepackage{pifont}       
\usepackage{bbding}       
\usepackage{fontawesome}  
\usepackage{tcolorbox}
\usepackage{soul}

\newcolumntype{P}[1]{>{\centering\arraybackslash}p{#1}}

\newlength{\Bheight}
\definecolor{codeblue}{rgb}{0.25,0.5,0.5}
\definecolor{myblue}{rgb}{0.88,0.98,1}
\definecolor{mygreen}{rgb}{0.92, 1.0, 0.92}
\definecolor{myred}{rgb}{1, 0.9, 0.9}
\definecolor{mygray}{gray}{0.95}
\newcommand{\colorrect}[1]{\textcolor{#1}{\ding{110}}}
\definecolor{Highlight}{HTML}{E8F8F5}
\definecolor{midgreen}{HTML}{69c5a3}
\definecolor{midblue}{HTML}{69a3f1}
\definecolor{darkgreen}{HTML}{146038}
\definecolor{darkblue}{HTML}{143b59}

\definecolor{hotpink}{RGB}{59, 115, 227}

\usepackage{amsthm}
\theoremstyle{plain}

\theoremstyle{definition}

\theoremstyle{remark}

\usepackage[textsize=tiny]{todonotes}

\newcommand{\seaemoji}{%
  \raisebox{-0.2em}{\includegraphics[scale=0.05]{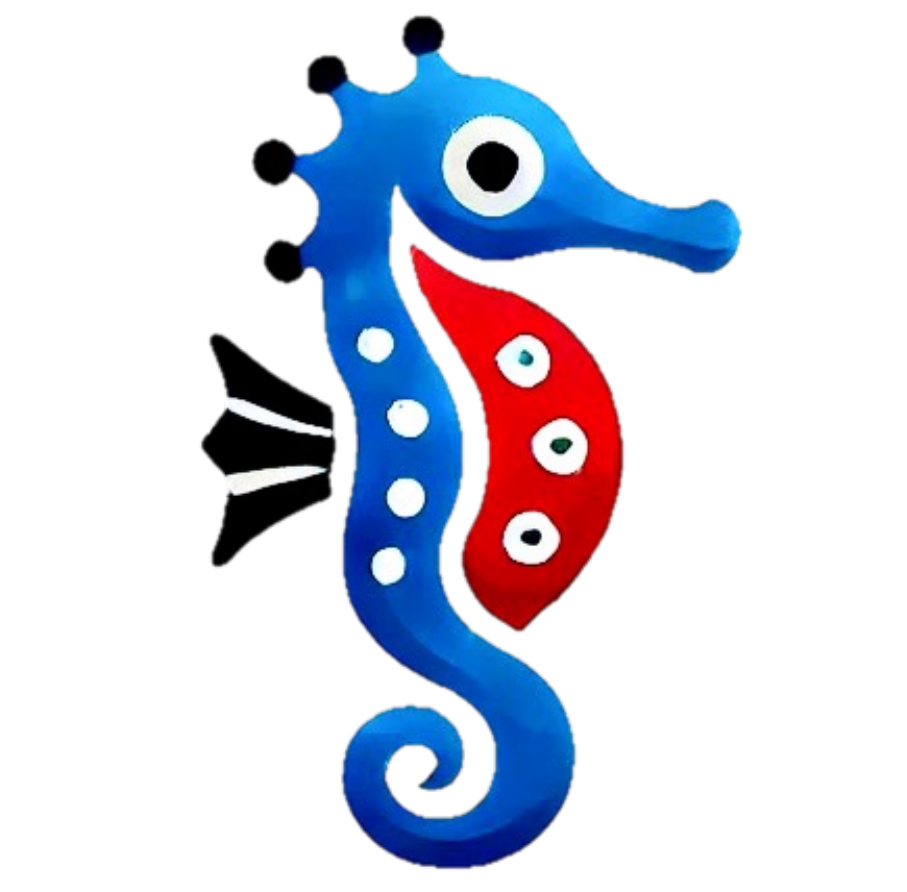}}%
}

\title{\centering \seaemoji{} SEA: Supervised Embedding Alignment for Token-Level Visual-Textual Integration in MLLMs}

\author{
  {\bf Yuanyang Yin}\thanks{~~Equal contribution.}$^{1}$ \quad
  {\bf Yaqi Zhao}\footnotemark[1]$^{2}$ \quad
  {\bf Yajie Zhang}$^{3}$ \quad
  {\bf Yuanxing Zhang}$^{3}$ \quad
  {\bf Ke Lin}$^{3}$ \quad
  {\bf Jiahao Wang}$^{3}$ \quad \\
  {\bf Xin Tao}$^{3}$ \quad
  {\bf Pengfei Wan}$^{3}$ \quad
  {\bf Wentao Zhang}\thanks{~~Corresponding author.}$^{2}$ \quad
  {\bf Feng Zhao}\footnotemark[2]$^{1}$ \\
  $^{1}$ University of Science and Technology of China \quad
  $^{2}$ Peking University \quad
  $^{3}$ Kuaishou Technology
}

\begin{document}
\maketitle

\begin{abstract}
Multimodal Large Language Models (MLLMs) have demonstrated remarkable capabilities by integrating visual and textual inputs, yet modality alignment remains one of the most challenging aspects. Current MLLMs typically rely on simple adapter architectures and pretraining approaches to bridge vision encoders with large language models (LLM), guided by
image-level supervision.
We identify this paradigm often leads to suboptimal alignment between modalities, significantly constraining the LLM's ability to properly interpret and reason with visual features particularly for smaller language models.
To address this fundamental limitation, we propose Supervised Embedding Alignment (SEA), a token-level supervision alignment method that enables more precise visual-text alignment during pretraining. SEA introduces minimal computational overhead while preserving language capabilities and substantially improving cross-modal understanding.
Our comprehensive analyses reveal critical insights into the adapter's role in multimodal integration, and extensive experiments demonstrate that SEA consistently improves performance across various model sizes, with smaller models benefiting the most (average performance gain of 7.61\% for Gemma-2B). This work establishes a foundation for developing more effective alignment strategies for future multimodal systems.

\end{abstract}

\section{Introduction}
Multimodal Large Language Models (MLLMs) have emerged as a transformative development in AI research, demonstrating exceptional capabilities in perceiving and reasoning across different modalities~\cite{agrawal2019nocaps,antol2015vqa,liu2023llava1.5,li2024llava,bai2025qwen2}. By integrating visual and textual information, these models mark a crucial step toward artificial general intelligence.

The standard MLLM pipeline consists of two-stages~\cite{liu2023llava1.5,liu2024llava,jiang2023mistral,minigpt4,instructblip,li2024llava,zhou2024tinyllava}: pre-training, where an adapter maps vision encoder features to the LLM's input space, guided by image-level supervision, and instruction tuning, which further adapts the model for downstream tasks, often involving partial or full LLM fine-tuning.

However, despite recent advances through scaling up data, models, and visual inputs~\cite{tong2024cambrian1fullyopenvisioncentric,li2024llava,bai2025qwen2,wang2024qwen2}, 
current approaches to cross-modal alignment in MLLMs predominantly rely on coarse-grained image-level or region-level supervision, such as optimal transport~\cite{park2024bridging} or regression-based techniques~\cite{shang2024llava}. These methods fail to capture the fine-grained semantics necessary for optimal visual-language integration. Therefore, the adapter's critical role of current alignment paradigm remain insufficiently explored.

\begin{figure*}[t!]
\centering
\subfigure[Accurate representation alignment with SEA.]{\label{fig:representation}
 \includegraphics[height=0.185\linewidth]{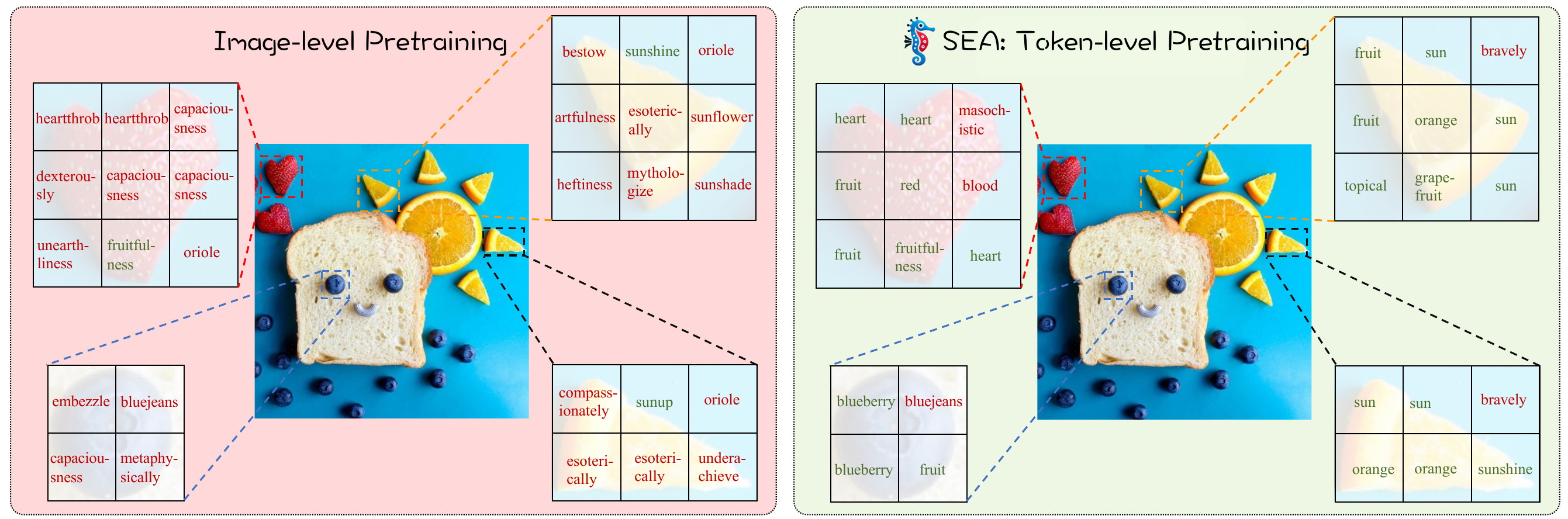}}
\hspace{0mm}
\subfigure[Improved visual perception enabled by SEA.]
{\label{fig:perception}
 \includegraphics[height=0.1905\linewidth]{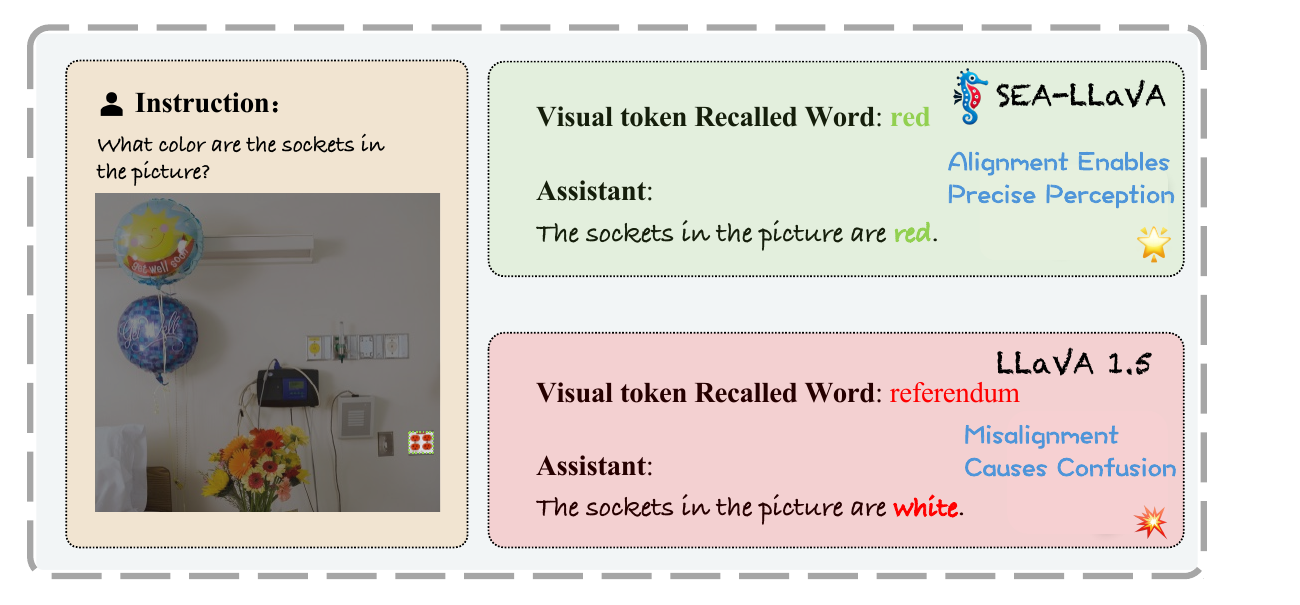}}
\caption{\footnotesize \textbf{Illustration of token-level alignment benefits}. (a) For each visual token, we retrieve and display the most similar word from the pre-defined vocabulary. SEA (right) produces semantically appropriate words (e.g., "blueberry", "orange") that better capture the visual content compared to conventional image-level alignment (left). (b) This improved alignment directly enhances visual perception capabilities, enabling more precise understanding of image elements (SEA-LLaVA correctly identifies "red" sockets while LLaVA misidentifies them).}
\label{fig:representation-perception}
\end{figure*}

\begin{figure}[t!]
    \centering
    \includegraphics[width=\linewidth]{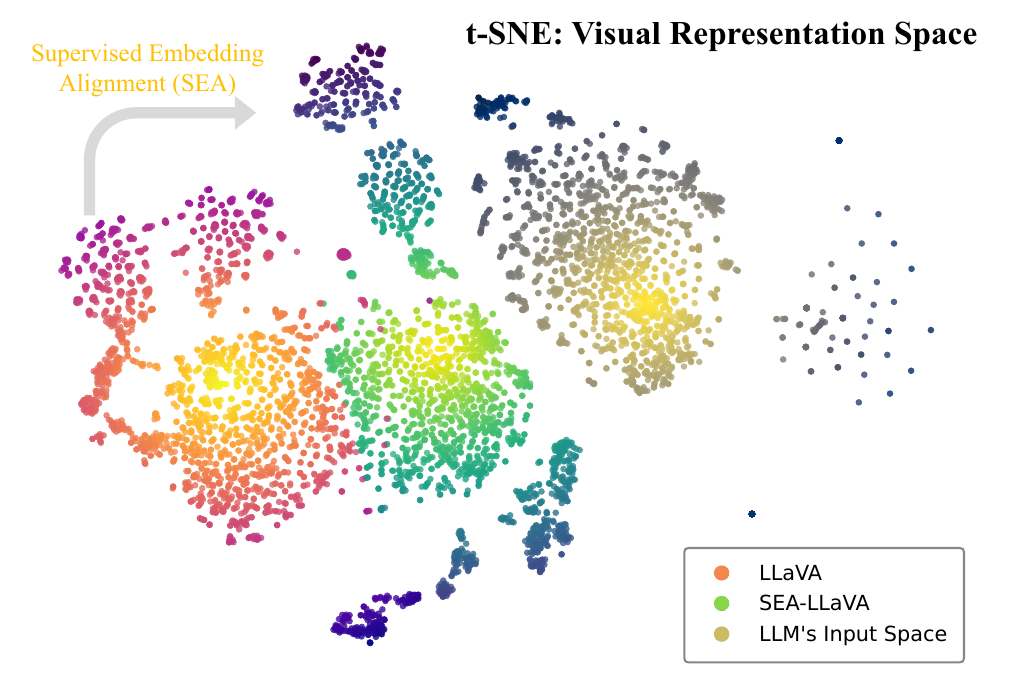}
     \caption{\footnotesize \textbf{Illustration of the distribution of different token embeddings.} Using t-SNE, we visualize the embedding space of LLaVA visual tokens (left), SEA-LLaVA visual tokens (mid), and LLM's native input embeddings (right). SEA effectively shifts visual token representations closer to the LLM's natural input space, reducing the adaptation burden on the language model and improving cross-modal integration.}
    \label{fig:t-SNE}
\end{figure}

Our experiments reveal two critical deficiencies in conventional image-level alignment. First, as shown in~\Cref{fig:representation-perception}, visual tokens from traditional adapters often fail to preserve their intended semantics, forcing the language model to compensate for these deficiencies and leading to incorrect visual understanding.
Second, and perhaps more concerning, the significant gap between adapter-processed visual tokens and the LLM's native input space (see~\Cref{fig:t-SNE}) requires the language model to allocate extra capacity interpreting misaligned visual inputs, rather than leveraging its pre-trained knowledge. 
These issues are particularly pronounced in smaller models, where limited capacity makes the trade-off between visual perception and language performance more severe.

This work addresses a fundamental question: \emph{How can we achieve optimal cross-modal alignment in MLLMs?} 
To effectively bridge the gap between modalities, we argue that alignment must occur at the token level, where individual visual tokens are precisely mapped to their corresponding semantic representations in the language space. However, achieving such fine-grained alignment presents fundamental challenges: visual tokens contain rich, multifaceted semantic information that cannot be trivially equated to single word tokens. Additionally, visual tokens often exhibit semantic shifts that cannot be easily captured through token-level annotations.  

To address this, we introduce \textbf{Supervised Embedding Alignment (SEA)}, which achieves optimal cross-modal alignment through token-level supervision during pretraining. By leveraging well-aligned vision-language models like CLIP, SEA obtains precise semantic labels for visual tokens and guides them toward optimal representations in the LLM's embedding space through contrastive learning (see~\Cref{fig:t-SNE}). This approach requires no additional training data or inference overhead.

Empirically, SEA demonstrates consistent improvements across model scales (2B-13B parameters), with particularly substantial gains for smaller models (7.61\% improvement on Gemma-2B). This scalability, combined with enhanced fine-grained visual perception, fundamentally addresses the limitations of current MLLM designs while maintaining computational efficiency.

In summary, our contributions and findings can be summarized as follows:
\begin{itemize}[leftmargin=*]
\setlength\itemsep{0.01em}
    \item We systematically analyze how adapter misalignment impacts MLLM performance, revealing its critical role in both visual perception and language capabilities.
    \item We propose SEA, a novel token-level alignment during pretraining 
    that effectively bridges the modality gap by precisely aligning visual tokens with the LLM's input space.
    \item We demonstrate SEA's effectiveness across model scales and different vision encoders without additional training data or inference overhead, showing particular benefits for smaller models.
\end{itemize}

\begin{figure}[t!]
    \centering     
        \subfigure[MMLU over Finetuning.]{\label{fig:mmlu}
        \includegraphics[width=0.48\linewidth]{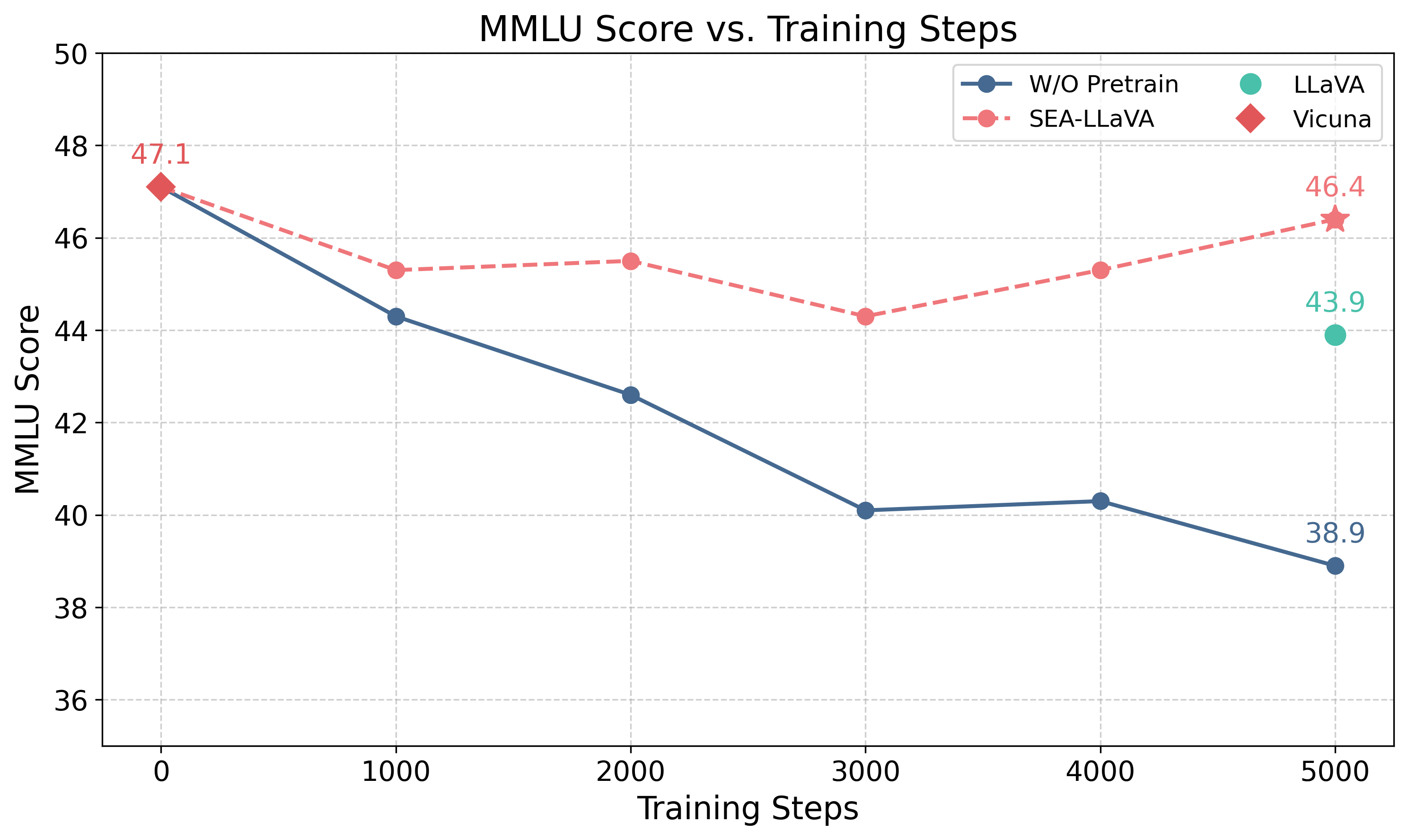}}
    \hspace{0mm}
        \subfigure[Results of Different Methods.]{\label{fig:radar}
        \includegraphics[width=0.373\linewidth]{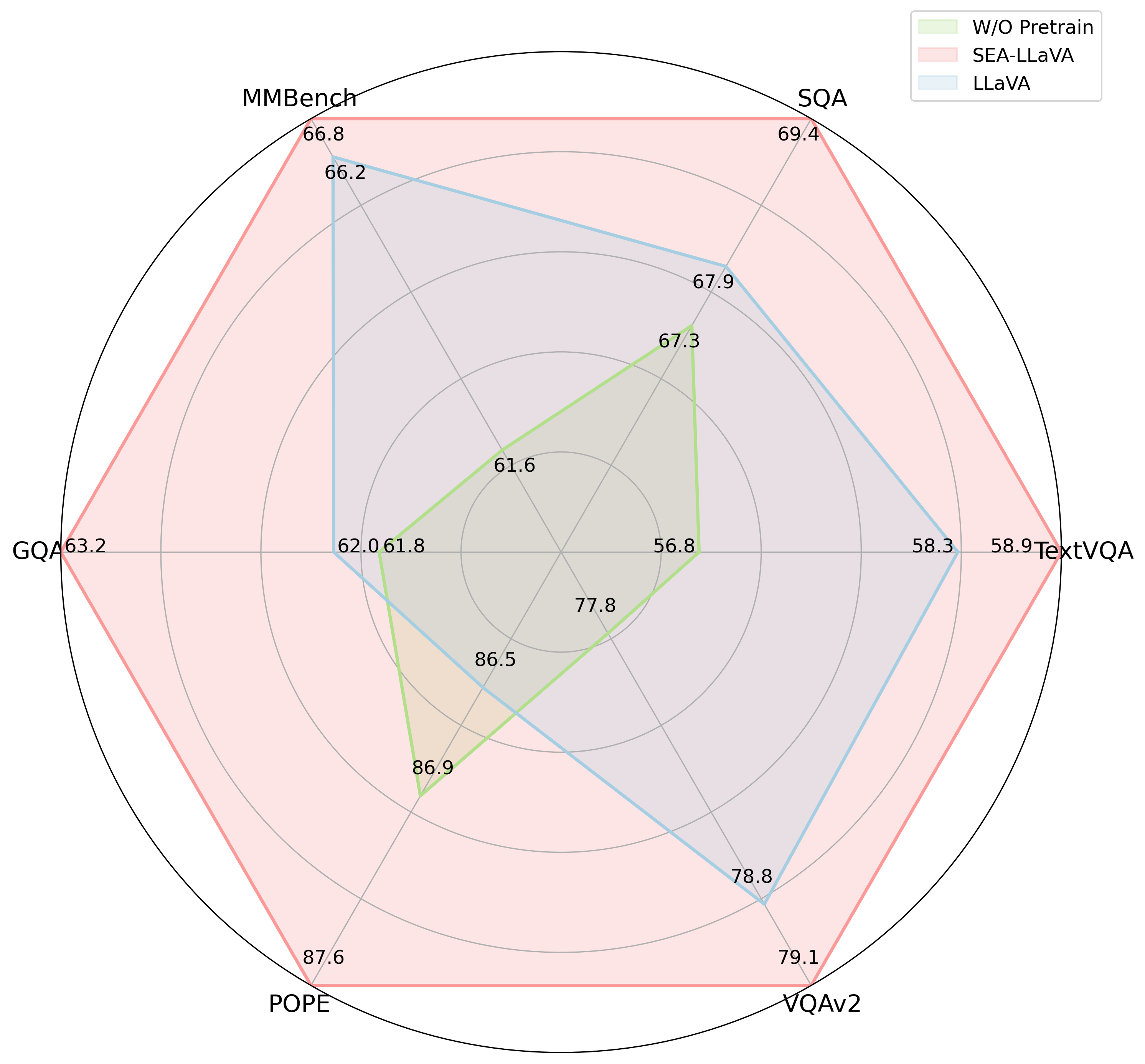}}
     \caption{\footnotesize \textbf{Impact of alignment quality on model performance}. (a) Language model capability (measured by MMLU score) during instruction-tuning: SEA-LLaVA (red line) maintains higher language capabilities compared to LLaVA (green point) by reducing adaptation burden. (b) Radar chart comparing performance across different benchmarks: SEA-LLaVA (red) consistently outperforms LLaVA (blue) on multiple evaluation metrics.}
    \label{fig:comparison}
\end{figure}
\section{Background and Problem Formulation}\label{sec:pre}
This section introduces the adapter-based architecture in MLLMs and analyzes the cross-modal misalignment problem that forms the foundation for our method in~\Cref{sec:method}.

\subsection{Adapter-Based Architecture in MLLMs}
Multimodal Large Language Models typically employ an adapter module to bridge vision encoders and language models. During pre-training, this adapter $g_{\theta}$ transforms visual patches output by the vision encoder $f$ into visual tokens compatible with the LLM's embedding space. 

For a given image-text pair $(X_{\text{image}},X_{\text{text}})$, the model processes inputs as follows:
\begin{equation}
    X_{v} = g_{\theta}(f(X_{\text{image}})) 
\end{equation}
\begin{equation}
    X_{t} = \Psi(X_{\text{text}})
\end{equation}
\begin{equation}
\begin{aligned}
    X_{\text{input}} &= [x_{v_0}, \dots, x_{v_m}, x_{t_0}, \dots, x_{t_n}] \\
    &\quad x_{v_j} \in X_v, \quad x_{t_i} \in X_t
\end{aligned}
\end{equation}
where $\Psi$ represents the LLM's embedding layer. The concatenated inputs $X_{\text{input}}$ are then processed by the LLM, with the adapter parameters $\theta$ updated using an auto-regressive language modeling loss.

\subsection{Issues in Image-level Alignment}
Despite current pre-training paradigm, significant misalignment issues persist between visual and textual representations in MLLMs. To quantitatively analyze this misalignment, we measure the semantic correspondence between visual tokens and language representations.

\paragraph{Semantic Information Distortion}
We evaluate the semantic information encoded in visual tokens by retrieving their closest word embeddings from a predefined word list $W$. For each visual token $x_{v_j} \in X_v$, we identify the word $w_j \in W$ with the highest similarity:
\begin{equation}
    w_j = \arg\max_{w \in W} \text{sim}(x_{v_j}, \Psi(w))
\end{equation}
where $\text{sim}(\cdot, \cdot)$ is the cosine similarity function. 
As shown in~\Cref{fig:representation}, conventional adapters frequently map visual tokens to semantically unrelated words (e.g., "bluejeans" for blueberries), indicating severe semantic distortion.
As shown in~\Cref{fig:perception}, this distortion forces the language model to compensate for representational discrepancies, resulting in incorrect visual understanding.

\paragraph{Modality Representation Gap}
We further analyze the modality gap through embedding space visualization (\Cref{fig:t-SNE}). 
We selected approximately 100 images from COCO val2014~\cite{chen2015microsoft} and generated detailed captions using Qwen2.5-VL~\cite{bai2025qwen2}. The visualization shows three distinct clusters: visual token embeddings ($X_v$) from the images (orange), text token embeddings from the captions (yellow). 
The significant distance between conventional adapter-processed visual tokens and text token embeddings reveals a fundamental representational gap. Mathematically, we can quantify this gap as:
\begin{equation}
    D = \frac{1}{|X_v|} \sum_{x_{v_j} \in X_v} \min_{w \in W} \left\| x_{v_j} - \Psi(w) \right\|_2
\end{equation}
This gap forces the language model to allocate substantial capacity to interpret misaligned visual inputs rather than leveraging its inherent knowledge.

The impact of this misalignment is clearly demonstrated in~\Cref{fig:comparison}, where we track the language model's performance (measured by MMLU score) during instruction-tuning. The model without pre-training (blue line) shows a substantial decrease in language capability as training progresses, highlighting the critical importance of alignment. However,the  conventional image-level alignment provides only marginal mitigation. This effect is particularly pronounced in smaller models where computational capacity is limited, highlighting the critical need for more efficient alignment strategies.

\begin{figure*}[t!]
    \centering
    \includegraphics[width=\linewidth]{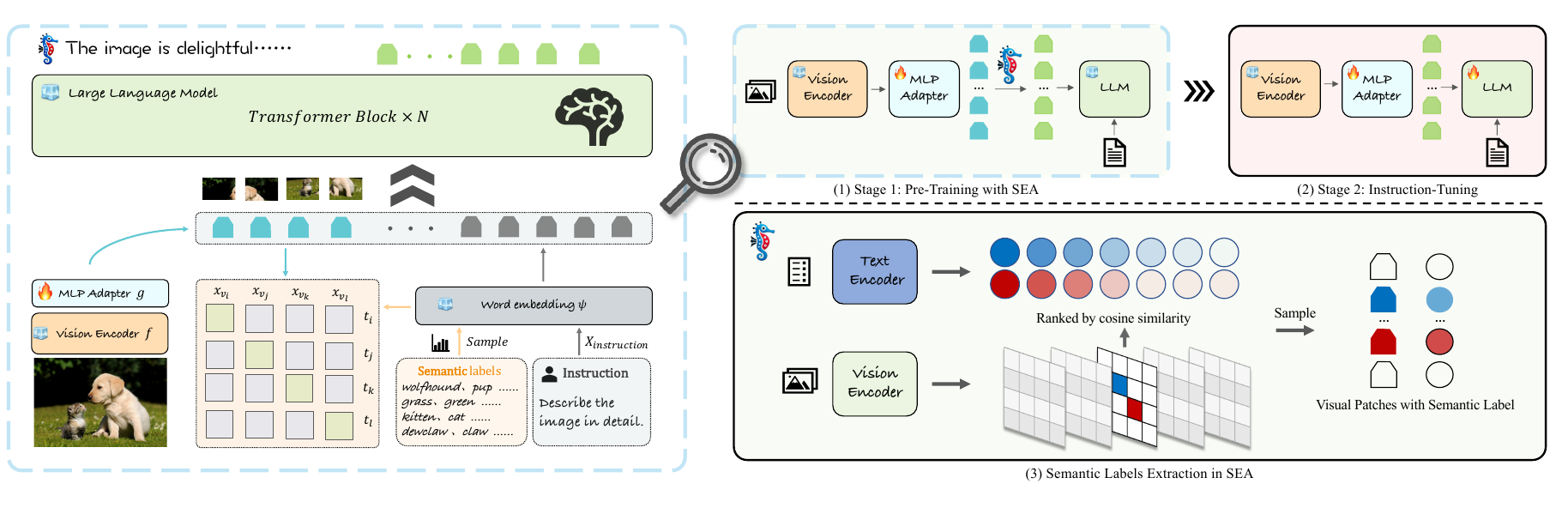}
    \caption{\footnotesize Left: \textbf{Overview of the proposed SEA.} For each visual token, SEA samples semantic labels with similarity-based weighting and identifies their corresponding representations in the LLM's embedding space. These are then used to supervise the adapter via contrastive learning, enabling token-level alignment. Right: \textbf{Overview of the SEA training pipeline.} During pretraining, SEA enhances modality alignment through token-level semantic supervision via contrastive learning, guided by candidate labels derived from the text encoder. Once alignment is established, visual tokens are mapped to representations more compatible with the LLM input space, substantially reducing the burden on the LLM during instruction tuning.}  
    \label{fig:method&pipeline}
\end{figure*}

\section{Method: Supervised Embedding Alignment}\label{sec:method}
This section presents SEA, the first supervision paradigm to mitigate the issue of misalignment between visual and text tokens in LLM's embedding space during pretraining (See~\Cref{fig:method&pipeline}). We will introduce each step of SEA in detail.

\subsection{Extract Semantic Labels for Visual Patches}

To achieve fine-grained supervision of the semantic feature expression for each visual token transformed by the adapter, we obtain continuous semantic labels for each patch after the vision encoder. For a pre-trained vision encoder $f$ paired with a text encoder $h$ and a word list $W$ containing $q$ words, we extract semantic information for each patch using Eqs.~\eqref{recall_1},~\eqref{recall_2},~\eqref{sim_score}, where $m$ is number of visual patches and $d$ is the dimention. We then select the top $n$ words based on cosine similarity scores for each visual patch (See~\Cref{fig:method&pipeline}(3)). To ensure only relevant and positively correlated words are considered, we exclude labels with similarity scores below 0. The remaining words serve as the \textit{semantic labels} for each visual patch. This approach assigns multiple semantic labels to each token, preserving its continuous semantic representation and preventing semantic shift through paired training of the vision and text encoders.

\begin{equation}
    V = f(X_{\text{image}})\in \mathcal{R}^{m\times d}
    \label{recall_1}
\end{equation}
\begin{equation}
    T = h(W)\in \mathcal{R}^{q\times d}
    \label{recall_2}
\end{equation}
\begin{equation}
w_i, s_i = \underset{j}{\mathrm{argmax}}\ \{-\mathrm{cos}(v_i, t_j)\}
\label{sim_score}
\end{equation}
where $w_i$ and $s_i$ are the indices and scores of the top $n$ semantic labels for the $i$-th visual patch $v_i$ respectively. $v_i$ is the visual feature of the patch obtained from the vision encoder $f$, and $t_j$ is the text embedding of the $j$-th word in the word list $W$, obtained from the text encoder $h$. The negative cosine similarity $\mathrm{-cos}(v_i, t_j)$ is computed as described in previous works~\cite{li2023clip}, where the cosine similarity needs to be negated in the CLIP embedding space.

\subsection{Token-Level Alignment}
The use of an adapter aims to convert visual patches into LLM's embedding space. However, the current image-level approach falls short of achieving this adequately as shown in \Cref{fig:representation}. We suggest using the semantic labels of each patch to directly guide the adapter in transforming visual patches into the LLM's embedding space, thereby reducing misalignment. 

\paragraph{Similarity-Weighted Sampling for Continuous Semantic Representation} 
Due to the semantic continuity of visual tokens, we should identify an appropriate position for each visual token within the LLM's embedding space, ensuring it retains its continuous semantic representation.
Specifically, for a given visual patch $v_{i}$ with its corresponding semantic labels $L_i=[w_1, \ldots, w_n]$ and similarity scores $S_i=[s_1, \ldots, s_n]$, we first normalize the similarity scores to get the sampling probability, and then sample a label for each patch based on $S^i_{norm}$ in Eq.~\eqref{norm}.
\begin{equation}
    S^i_{norm}= \frac{S_i}{sum(S_i)}
    \label{norm}
\end{equation}

\paragraph{A Localized Sampling Strategy}
To further enhance the effectiveness of contrastive learning and mitigate the issue of excessive similarity between samples, we adopt a localized sampling strategy. For each image, we perform sampling within a $k \times k$ window, ensuring that only one patch is sampled from each window. Consequently, a single image with $N$ visual patches will have $N / (k \times k)$ patches participating in contrastive learning. For visual patches sharing the same label in one batch, we randomly retain only one patch to ensure the effectiveness of contrastive learning. We then obtain a series of visual patches with labels, namely, $\{(x_{v_1},w_1),\ldots,(x_{v_N},w_N)\}$, where $N$ is the number of tokens in one batch. 

For each label $w_i$, we compute the corresponding text feature $t_i$ as follows:
\begin{equation}
    t_i = \frac{1}{M} \sum_{k=1}^M \Psi(w_i^k)
\end{equation}
where $\Psi(w_i^k)$ represents encoded feature of the $k$-th token of $w_i$, and $M$ is the number of tokens after encoding $w_i$.

The loss of alignment can be computed as:
\begin{equation}
\begin{split}
    \mathcal{L}_a = -\frac{1}{2N} \sum_{i=1}^N \Bigg( & \log \frac{\exp(\phi(\boldsymbol{x_v}_i, \boldsymbol{t}_i) / \tau)}{\sum_{j=1}^N \exp(\phi(\boldsymbol{x_v}_i, \boldsymbol{t}_j) / \tau)} \\
    & + \log \frac{\exp(\phi(\boldsymbol{t}_i, \boldsymbol{x_v}_i) / \tau)}{\sum_{j=1}^N \exp(\phi(\boldsymbol{t}_i, \boldsymbol{x_v}_j) / \tau)} \Bigg)
\end{split}
\end{equation}
where $\phi({\boldsymbol{x_v}}_i,{\boldsymbol{t}}_j)=\frac{{\boldsymbol{x_v}}_i}{\|{\boldsymbol{x_v}}_i\|_2}\cdot\frac{{\boldsymbol{t}}_j}{\|{\boldsymbol{t}}_j\|_2}$, and $\tau$ is the temperature, a learnable parameter.

For generation, the prediction of the next token $x^{(i)}$ is conducted based on visual tokens $V_{i}$, prompt $P$ and previous tokens $x^{(<i)}$. The loss can be computed as:
\begin{equation}
\mathcal{L}_{\mathrm{g}} = -\frac{1}{B} \sum_{i=1}^{B} \log p_{\theta}\left(x^{(i)} \mid V_{\text{i}}, \mathrm{P}, x^{(<i)}\right)
\end{equation}
where $B$ is the batch size, $\theta$ is the trainable parameters.

During the pretraining process, two learning objectives simultaneously supervise the adapter. We obtain the final loss $\mathcal{L}$ of pretraining by adding $\mathcal{L}_a$ and $\mathcal{L}_g$, a weighting factor $\lambda$ is introduced to balance the two losses.
\begin{equation}
\mathcal{L} = \mathcal{L}_{\mathrm{g}} + \lambda \mathcal{L}_{\mathrm{a}}
\end{equation}
\section{Experiments}\label{sec:exp}
In this section, we conduct comprehensive experiments to validate SEA's effectiveness. First, we provide our evaluation results on 8 common benchmarks compared with different backbones. Then, we analyze how SEA enhances token-level alignment, visual perception and language capability. Finally, we explore SEA's generalization capability through extensive ablation studies.

\subsection{Experimental Setup}
We evaluate SEA's generalization capability across different MLLM components:
1) Vision Encoders: We experiment with widely-adopted vision encoders including CLIP-ViT-L@336px~\cite{CLIP} and SigLIP-ViT-SO@384px~\cite{SigLIP}.
2) Language Models: To assess scalability, we test SEA on LLMs ranging from 2B to 13B parameters, including Gemma-2B~\cite{gemma}, Phi-3-mini-4k-instruct~\cite{abdin2024phi3technicalreporthighly}, Llama3-8B-Instruct~\cite{llama3modelcard}, and Vicuna-1.5-7B\&13B~\cite{vicuna}.
3) SEA Configuration: We employ top-10 semantic labels ($n=10$), zero temperature ($\tau=0$) for robust alignment, and $2\times2$ window sampling for efficient training.
Additional training details and datasets are provided in~\Cref{appendix:exp}.

\begin{table*}[h!]
\centering
\resizebox{\linewidth}{!}{
\begin{tabular}{ccc|cccccccc}
\toprule
Method & LLM & Res. & {\bf VQA}$^\text{v2}$ & {\bf VQA}$^\text{T}$ &{\bf GQA} & {\bf SQA}$^\text{I}$ & {\bf MMB}  & {\bf POPE} & {\bf VizWiz} & {\bf MM-Vet}\\
\midrule
MobileVLM-3B\cite{chu2023mobilevlmfaststrong} & MLLaMA 2.7B & 336 & --  & 47.5 & 59.0 & 61.0 & 59.6 & 84.9 & --  & --\\
MobileVLM-V2-3B\cite{chu2024mobilevlmv2fasterstronger} & MLLaMA 2.7B & 336 & --  & 57.5 & 61.1  & 70.0 & 63.2 & 84.7 & --  & --\\
LLaVA-Phi \cite{zhu2024llavaphiefficientmultimodalassistant} & Phi-2.7B & 336 & 71.4  & 48.6 & --  & 68.4 & 59.8 & 85.0 & 35.9 &28.7  \\
TinyLLaVA \cite{zhou2024tinyllavaframeworksmallscalelarge} & Phi-2.7B & 384 & 79.9  & 59.1 & 62.0  & 69.1 & 66.9 & 86.4 & -- &32.0  \\
InstructBLIP~\cite{instructblip} & Vicuna-7B & 224 & -- & 50.1 & -- & -- & 30.6  & -- &34.5  & -- \\
InstructBLIP~\cite{instructblip} & Vicuna-13B & 224 & --  & 50.7 & 49.5 & 63.1 & --  & --  &33.4  & --  \\
Qwen-VL~\cite{bai2023qwen} & Qwen-7B & 448 & 79.5  & 63.8 & 59.3 & 67.1 & 38.2  & -- &35.2 &--\\
Qwen-VL-Chat~\cite{bai2023qwen} & Qwen-7B & 448 & 78.2  & 61.5 & 57.5  & 68.2 & 60.6  & -- &38.9 &--  \\
LLaMA-VID~\cite{llamavid} & Vicuna-7B & 336 & 79.3  & -- & 64.3  & 68.3 & 65.1 & 86.0 & 54.2 & -- \\
LLaMA-VID~\cite{llamavid} & Vicuna-13B & 336 & 80.0  & -- & \underline{65.0}  & 70.0 & 66.6  &  86.0  & 54.3 & --\\
LLaVA-1.5$^{*}$~\cite{liu2023llava1.5} & Vicuna-7B & 336 & 78.8  & 58.3 & 62.0  & 67.9 & 66.2  & 86.5 & 45.7 &30.7  \\
LLaVA-1.5$^{*}$~\cite{liu2023llava1.5} & Vicuna-13B & 336 & 80.0  & 60.8 & 63.3  & 71.6 & 67.7  & \underline{87.6} & 53.6 &35.1  \\
ShareGPT4V~\cite{chen2023sharegpt4vimprovinglargemultimodal}& Vicuna-7B & 336 & 80.6 & -- & --  & 68.4 & 68.8  & -- & -- &37.6 \\
{Mini-Gemini}~\cite{Mini-Gemini} & Gemma-2B & 336+768 & -- & 56.2 & --  & -- & 59.8  & --  & -- & 31.1\\
{Mini-Gemini}~\cite{Mini-Gemini} & Vicuna-7B & 336+768 & -- & 65.2 & --  & -- & 69.3  & --  & -- & 40.8\\
{Mini-Gemini}~\cite{Mini-Gemini} & Vicuna-13B & 336+768 & -- & 65.9 & --  & -- & 68.5  & --  & -- &46.0\\
S$^2-$Wrapper$^{*}$~\cite{shi2024need} & Vicuna-7B  & 1008 & 79.7  & 60.3 & 63.2  & -- & 67.3   & 87.4  & 50.1 &33.0\\
S$^2-$Wrapper~\cite{shi2024need} & Vicuna-13B  & 1008 & 80.9  & 63.1 & --  & -- & 67.9  & --  & 56.0 &35.4\\
AlignGPT~\cite{zhao2024aligngpt} & Vicuna-7B & 336 & 79.1  & 58.4 & 62.9 & 68.5 & 67.3  & 86.0  & 54.2 & 30.8\\
AlignGPT~\cite{zhao2024aligngpt} & Vicuna-13B & 336 & 80.0 & 60.2 & 63.6 & 70.3 & 69.5  & 86.2  & 56.4 & 35.6 \\
Visual Prompt~\cite{lin2024rethinking} & Vicuna-7B & 336 & 79.8 & 59.8 & 63.3 & 69.5 & 67.6  & {\bf88.9}  & --  &34.9\\
\midrule\midrule
\multicolumn{10}{c}{\small \em Our Models}\\ \midrule
\rowcolor{mygreen}
{\bf SEA-PRIME} & Gemma-2B & 384 & 81.0 & 60.7& 62.4 & 69.2 & 68.8 & 87.8 & 61.9 & 38.0 \\
\rowcolor{mygreen}
{\bf SEA-PRIME} & Phi3-3.8B & 384 & 80.7 & 64.0& 62.0 & {78.7} & 72.6 & 87.0 & 61.9 & \underline{46.8} \\
\rowcolor{mygreen}
{\bf SEA-PRIME} & Vicuna-7B & 384 & 81.4 & \underline{67.2} & {63.1} & {73.9} & {75.6}  & \underline{88.4}  & \underline{63.8} &44.2\\
\rowcolor{mygreen}
{\bf SEA-PRIME} & Llama3-8B & 384 & {\bf83.1} & {\bf68.0} & {\bf65.1}  & \underline{79.0} & \underline{76.0}  & {87.4}  & \bf{64.7} & {46.0}\\
\rowcolor{mygreen}
{\bf SEA-PRIME} & Vicuna-13B & 384 & \underline{81.9}& {66.2} & 64.3 &{\bf80.9} & {\bf 76.9}   &  86.7 & {63.6} & {\bf {48.8}}\\
\bottomrule
\end{tabular}
}
\small{
\centering
\caption{{\footnotesize \label{tab:main_result} \bf Main evaluation results compared with leading baselines on 8 popular benchmarks.} VQA$^\text{v2}$~\cite{vqav2}; VQA$^\text{T}$: TextVQA~\cite{textvqa}; GQA~\cite{hudson2019gqa}; SQA$^\text{I} $:ScienceQA-IMG~\cite{lu2022scienceqa}; MMB: MMBench~\cite{liu2023mmbench}; POPE~\cite{li2023pope}; VizWiz~\cite{gurari2018vizwizgrandchallengeanswering}; MM-Vet~\cite{yu2023mmvet}. All methods maintain the number of visual tokens without doubling, and models marked with * are results we reproduced. Column Res. is the image resolution of vision model.}
}
\end{table*}

\subsection{Main Results}
We leverage SEA to train a family of MLLMs called SEA-PRIME, utilizing LLM backbones of various scales. The vision component employs SigLIP-ViT-SO400M/14@384. We pre-train the connector using 2.5M adapter data and instruction tune using Cambrian-7M~\cite{tong2024cambrian1fullyopenvisioncentric}.

As shown in~\Cref{tab:main_result}, SEA-PRIME show robust improvements over existing open-source methods. 
Even with smaller models (2B and 3.8B), it achieves competitive results compared to larger counterparts. The scalability becomes particularly evident with LLaMA-3-Instruct-8B~\cite{llama3modelcard}, where SEA-PRIME demonstrates superior performance across all benchmarks.

These results highlight SEA's ability to enhance model performance while maintaining efficiency, particularly benefiting smaller models through better alignment.
\begin{figure}[!t]
    \centering
    \includegraphics[width=0.8\linewidth]{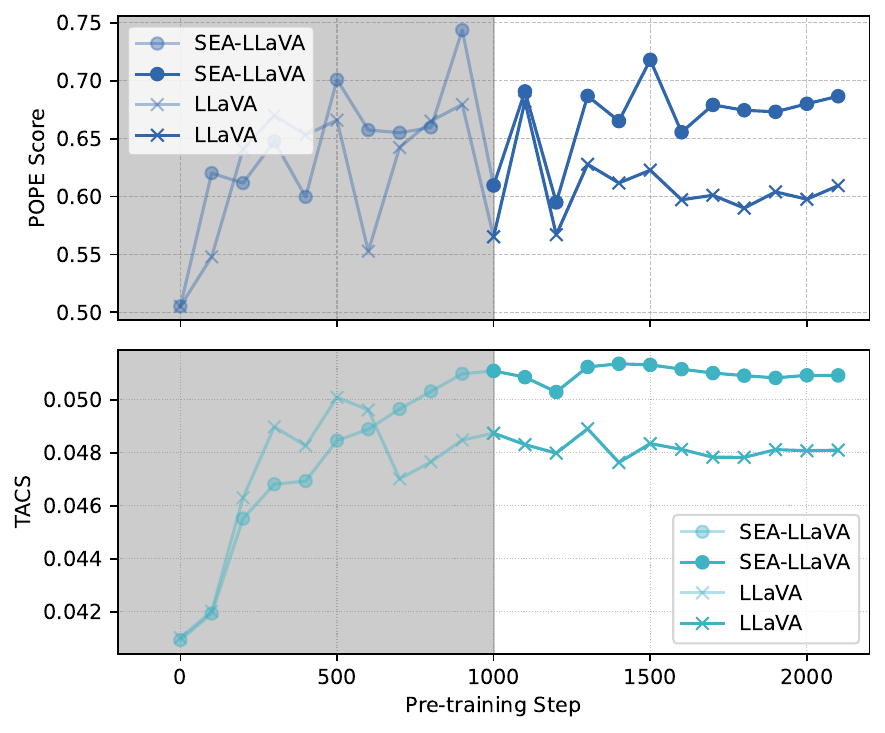}
    \caption{\footnotesize {\bf TACS score and POPE score during pre-training.} SEA achieves better alignment and higher POPE scores under the same training data.}  
    \label{fig:Aligment-POPE}
\end{figure}

\subsection{Token-level Alignment Analysis}
To comprehensively evaluate SEA's effectiveness in bridging the modality gap, we analyze its impact from three perspectives: alignment quality during pre-training, fine-grained visual perception, and preservation of language capabilities.

\paragraph{Alignment Quality} To quantify alignment quality, we introduce Token Alignment Consistency Score (TACS) (see~\Cref{appendix:alignment_score}), a simple metric aligned with the SEA loss. As shown in~\Cref{fig:Aligment-POPE}, SEA progressively improves TACS during training, with corresponding gains in POPE scores. This correlation validates both our metric and SEA's effectiveness in improving visual-text integration.

\paragraph{Fine-grained Visual Perception} As illustrated in~\Cref{sec:pre}, conventional MLLMs treat visual tokens as additional vocabulary, limiting their semantic understanding (see~\Cref{fig:representation}). SEA addresses this by providing precise semantic supervision during pre-training, enabling more accurate visual token representations. This improvement in token-level alignment directly enhances the model's ability to capture fine-grained visual semantics, as demonstrated across diverse perception-focused tasks (see~\Cref{tab:perception-bench}). From detailed caption generation to fine-grained object recognition, SEA consistently improves the model's visual understanding capabilities.

\paragraph{Language Model Capabilities} 
A key challenge in multimodal learning is maintaining the LLM's inherent language abilities while adapting to visual inputs. As shown in \Cref{fig:mmlu}, conventional image-level alignment show degradation (green point) in language performance after training. In contrast, SEA's semantically aligned visual representations alleviate the adaption burden, allowing the language model to better preserve its pretrained knowledge and capabilities.

These analyses demonstrate that SEA effectively address both semantic distortion and modality representation gaps identifies in~\Cref{sec:pre}, leading to improved overall model performance.

\begin{table*}[!t]
\small{
  \centering
  \label{tab:ablation}
  \resizebox{\linewidth}{!}{
  \begin{tabular}{c c c c c | cccccccc }
    \toprule
    Method & VE & Res. & PT+IT & LLM & {\bf VQA}$^\text{v2}$ & {\bf VQA}$^\text{T}$ &{\bf GQA} & {\bf SQA}$^\text{I}$ & {\bf MMB} & {\bf POPE} & {\bf VizWiz }  & {\bf MM-Vet }\\
    \midrule
    LLaVA & CLIP-L & 336 & 0.5M+0.6M & Vicuna-7B & 78.8  & 58.3 & 62.0  & 67.9 & 66.2  & 86.5 & 45.7 & 30.7 \\
    \cellcolor{mygreen}  {SEA-LLaVA} & CLIP-L & 336 & 0.5M+0.6M & Vicuna-7B  & {\bf 79.1}  & {\bf 58.9} &  {\bf 63.2} & \bf69.4 & {\bf 66.8}   & {\bf 87.6}  & {\bf 48.8}& {\bf31.9} \\

    \midrule \midrule
    \multicolumn{12}{c}{\small \em Applying to Different LLMs}\\ \midrule
    LLaVA & CLIP-L & 336 & 0.5M+0.6M & Gemma-2B & 72.5  & 43.7 & 56.0  & 61.3 & 54.0  & 84.4  & 38.7& 23.9 \\
    \cellcolor{mygreen} {\em + SEA} & CLIP-L & 336 & 0.5M+0.6M & Gemma-2B  & {\bf 76.6}  & {\bf 49.7} & {\bf 60.9}  & {\bf 62.5} & {\bf 59.5}  & {\bf 87.0} &  {\bf39.5}& {\bf27.6}  \\ \midrule
    LLaVA & CLIP-L & 336 & 0.5M+0.6M & Phi3-3.8B  & 77.4  & 54.6 & 60.8  & 73.0 & 68.7  & 86.5 & 37.1 & \bf35.4\\
    \cellcolor{mygreen} {\em + SEA} & CLIP-L & 336 & 0.5M+0.6M & Phi3-3.8B  & {\bf 77.5}  & {\bf 55.3} & {\bf 61.0}  & {\bf 74.2} & {\bf 69.4}  & {\bf 87.0} &  {\bf39.0} & 34.7 \\ \midrule
    LLaVA & CLIP-L & 336 & 0.5M+0.6M & LlaMA3-8B & 79.4  & 57.7 & 63.7  & 76.0 & {\bf 72.5} & 87.0  & {\bf48.1}&{34.0} \\
    \cellcolor{mygreen}  {\em + SEA} & CLIP-L & 336 & 0.5M+0.6M & LlaMA3-8B  & {\bf 79.6}  & {\bf 58.0} & {\bf 63.8}  & {\bf 76.6} & 72.0  & 87.0 &45.2 &{\bf 36.3} \\
    \midrule
    LLaVA & CLIP-L & 336 & 0.5M+0.6M & Vicuna-13B & \bf 80.0  & {\bf 60.8} & 63.3  & 71.6 & 67.7 & { 87.6}  & {53.6}& {35.1}  \\
    \cellcolor{mygreen}  {\em + SEA} & CLIP-L & 336 & 0.5M+0.6M & Vicuna-13B  & 79.8  & 60.4 & {\bf 63.8}  & {\bf 71.7} & {\bf 68.0}  & 87.6 &{\bf 57.3}& \bf 35.8   \\\midrule\midrule
    \multicolumn{12}{c}{\small \em Applying to Different Vision Encoders}\\ \midrule
    LLaVA & SigLIP-SO & 384 & 0.5M+0.6M & Vicuna-7B & 80.8  & 62.3 & 63.2  & 70.6 & 68.0  & 86.7 & 51.1& 32.9  \\
    \cellcolor{mygreen}  {\em + SEA} & SigLIP-SO & 384 & 0.5M+0.6M & Vicuna-7B & {\bf 80.9}  & {\bf 62.6} & {\bf 63.4}  & {\bf 71.3} & {\bf 68.4}   & {\bf 87.3} & {\bf52.4}& \bf34.6   \\
    \bottomrule
  \end{tabular}}
  \caption{\footnotesize {\bf Exploring the compatibility and scalability of SEA.} Scaling results on LLM, vision encoder (VE) and resolution (Res.) are provided. "0.5M+0.6M" denotes the training data from LLaVA-1.5. Results with SEA are marked in \colorrect{mygreen}.}
  }
\end{table*}

\subsection{Ablation Study}
We conducted a comprehensive ablation study to evaluate the effectiveness of SEA. As shown in~\Cref{tab:ablation}, SEA introduces no additional training data or inference cost, yet consistently improves the overall performance of MLLMs. 

\vspace{2mm}
\noindent\textbf{SEA consistently benefits different LLMs, with particularly strong improvements in smaller models.}
Our experiments explore the application of SEA across LLMs of varying sizes. Notably, for the smaller model, SEA significantly boosts performance across multiple tasks, with an average performance gain of 7.61\%.
This highlights SEA's ability to effectively address misalignment issues that are more pronounced in smaller LLMs, thereby enhancing their performance. Larger LLMs, while inherently better at handling misalignment, still benefit from SEA, indicating that SEA offers additional alignment gains regardless of model size.

\vspace{2mm}
\noindent\textbf{SEA provides robust benefits across diverse vision encoders.}
We also examined the impact of SEA with different vision encoders. Replacing the CLIP-ViT~\cite{CLIP} with the SigLIP-SO(400M)~\cite{SigLIP}, SEA consistently improves performance, underscoring SEA's robustness across different encoders. 


\begin{table*}[!t]
  \centering
  \small
  \setlength{\tabcolsep}{6pt}
  \resizebox{\linewidth}{!}{
  \begin{tabular}{l | c r@{\hspace{3pt}}l | c r@{\hspace{3pt}}l | c r@{\hspace{3pt}}l | c r@{\hspace{3pt}}l | c r@{\hspace{3pt}}l}
    \toprule
    Method & \multicolumn{3}{c|}{\bf CapsBench} & \multicolumn{3}{c|}{\bf Stanford Dogs} & \multicolumn{3}{c|}{\bf COCO Captions (CIDEr)} & \multicolumn{3}{c|}{\bf OCRBench} & \multicolumn{3}{c}{\bf MMMU} \\
    \midrule
    LLaVA & 88.0 & & & 28.6 & & & 84.8 & & & 319 & & & 0.44 & & \\
    \cellcolor{mygreen}{\em + SEA} & 
{\bf 90.4} & \textcolor{darkgreen}{(+2.7\%)} & \textcolor{midgreen}{\rule{2.9mm}{2mm}} &
{\bf 29.7} & \textcolor{darkgreen}{(+3.9\%)} & \textcolor{midgreen}{\rule{4.2mm}{2mm}} &
{\bf 88.7} & \textcolor{darkgreen}{(+4.6\%)} & \textcolor{midgreen}{\rule{5.0mm}{2mm}} &
{\bf 336} & \textcolor{darkgreen}{(+5.3\%)} & \textcolor{midgreen}{\rule{5.5mm}{2mm}} &
{\bf 0.49} & \textcolor{darkgreen}{(+11.4\%)} & \textcolor{midgreen}{\rule{11.8mm}{2mm}} \\

    \bottomrule
  \end{tabular}
  }
  \caption{\footnotesize \textbf{Ablation results on fine-grained perception tasks.} We conduct ablation studies based on LLaVA across five fine-grained benchmarks: CapsBench~\cite{liu2024playground}, Stanford Dogs~\cite{dataset2011novel}, COCO Captions~\cite{chen2015microsoft}, OCRBench~\cite{liu2024ocrbench}, and MMMU~\cite{yue2024mmmu}. For Stanford Dogs, we reformulate the task as a 4-way multiple-choice question. Results show that SEA consistently improves the perceptual capabilities of MLLMs, particularly in capturing fine-grained visual semantics.}
  \label{tab:perception-bench}
  \vspace{-1mm}
\end{table*}

\begin{table*}[!t]
\small{
\label{tab:abla_ve}
\setlength{\tabcolsep}{4.3pt}
\centering
\resizebox{\linewidth}{!}{
\begin{tabular}{l | c r@{\hspace{3pt}}l | c r@{\hspace{3pt}}l | c r@{\hspace{3pt}}l | c r@{\hspace{3pt}}l | c r@{\hspace{3pt}}l | c r@{\hspace{3pt}}l }
  \toprule
  Method  & \multicolumn{3}{c|}{\bf VQA$^\text{v2}$} & \multicolumn{3}{c|}{\bf VQA$^\text{T}$} & \multicolumn{3}{c|}{\bf GQA} & \multicolumn{3}{c|}{\bf SQA} & \multicolumn{3}{c|}{\bf MMB} & \multicolumn{3}{c}{\bf VizWiz} \\
  \midrule
  Baseline  & 78.8 & & & 58.3 & & & 62.0 & & & 67.9 & & & 66.2 & & & 45.7 & & \\
  \cellcolor{mygreen}
  {\em + Finetune VE}   & 80.3& \textcolor{darkgreen}{+1.5} & \textcolor{midgreen}{\rule{5mm}{2mm}} & 59.1 & \textcolor{darkgreen}{+0.8}& \textcolor{midgreen}{\rule{2.67mm}{2mm}} & 63.4 & \textcolor{darkgreen}{+1.4} & \textcolor{midgreen}{\rule{4.67mm}{2mm}} & 67.0 & \textcolor{darkblue}{-0.9} & \textcolor{midblue}{\rule{3mm}{2mm}} & 66.1 & \textcolor{darkblue}{-0.1} & \textcolor{midblue}{\rule{0.33mm}{2mm}} & 50.3 & \textcolor{darkgreen}{+4.6} & \textcolor{midgreen}{\rule{8mm}{2mm}} \\ 
  \cellcolor{mygreen}
  {\em + SEA}   & {\bf 80.5} & \textcolor{darkgreen}{+0.2} & \textcolor{midgreen}{\rule{0.67mm}{2mm}} & {\bf 59.5} & \textcolor{darkgreen}{+0.4} & \textcolor{midgreen}{\rule{1.33mm}{2mm}} & {\bf 63.6} & \textcolor{darkgreen}{+0.2} & \textcolor{midgreen}{\rule{0.67mm}{2mm}} & {\bf 69.5} & \textcolor{darkgreen}{+2.5} & \textcolor{midgreen}{\rule{8.33mm}{2mm}} & {\bf 68.0} & \textcolor{darkgreen}{+1.9} & \textcolor{midgreen}{\rule{6.33mm}{2mm}} & {\bf 51.6} & \textcolor{darkgreen}{+1.3} & \textcolor{midgreen}{\rule{2.27mm}{2mm}}\\
  \bottomrule
\end{tabular}
}
\caption{{\footnotesize \bf Ablations for fine-tuning vision encoder.} The baseline is LLaVA-1.5 with Vicuna-7B, using the same training data and strategy. "Finetune VE" refers to the vision encoder is unfrozen during instruction tuning.}
}
\end{table*}

\begin{table*}[!t]
  \resizebox{\linewidth}{!}{
  \begin{tabular}{c c c c c | ccccccccc }
    \toprule
    Method & VE & Res. & PT+IT & LLM & {\bf VQA}$^\text{v2}$ & {\bf VQA}$^\text{T}$ &{\bf GQA} & {\bf SQA}$^\text{I}$ & {\bf MMB} & {\bf POPE} & {\bf VizWiz }  & {\bf MM-Vet } & {\bf MMVP }\\
    \midrule
    LLaVA & DINOv2-L & 224 & 0.5M+0.6M & Vicuna-7B & 71.4  & {45.8} & 58.6  & 63.9 & 54.2  & 84.8 & 37.6 & {\bf20.9} & 31.3 \\
     \cellcolor{mygreen}{\em + SEA} & DINOv2-L & 224 & 0.5M+0.6M & Vicuna-7B  & {\bf 74.0}  & {45.8} &  {\bf 60.9} & \bf65.1 & {\bf 57.6}   & {\bf 86.1}  & {\bf 39.6}& {20.8} & {\bf32.0} \\
    \bottomrule
  \end{tabular}
  }
  \small{
  \centering
  \caption{{\footnotesize \bf Exploring the semantic label transfer.} We obtained semantic labels from CLIP-Large and directly transferred them to the training of DINOv2, resulting in significant performance improvements.}
  \label{tab:Transfer}
}
\end{table*}

\subsection{Further Discussions}
\paragraph{Vision Encoder Fine-tuning}
Given that SEA leverages well-aligned vision encoders for optimal token-level supervision during pretraining, a natural concern arises: would fine-tuning the vision encoder in instruction-tuning potentially disrupt this carefully established alignment? To investigate this, we follow~\cite{tong2024cambrian1fullyopenvisioncentric}  to unfreeze the vision encoder during instruction-tuning. Surprisingly, our results show that this not only maintains but further improves performance (see~\Cref{tab:abla_ve}). This suggests that with SEA's strong token-level alignment as initialization, the vision encoder can focus on adapting to domain-specific features while preserving the semantic alignment established in pretraining.  These findings indicate SEA's flexibility and adaptability in different training paradigms.
\paragraph{Cross-encoder Transfer}
Recent advances in combining different vision encoders have shown promising results in MLLMs~\cite{tong2024eyes,tong2024cambrian1fullyopenvisioncentric,Mini-Gemini,goncharova2024omnifusion}, yet a common challenge lies in endowing these task-specific vision encoders with rich semantic understanding. We explore whether SEA's semantic supervision can bridge this gap by transferring CLIP-derived semantic labels to other vision encoders. Specifically, we apply SEA's training paradigm to DINOv2~\cite{oquab2023dinov2}, using the same semantic labels extracted from CLIP. As shown in~\Cref{tab:Transfer}, this simple transfer strategy leads to significant improvements on visual understanding benchmarks (e.g., VQAv2, GQA). Notably, the performance gains persist even on MMVP~\cite{tong2024eyes}, where DINOv2 traditionally excels. These results demonstrate that SEA's semantic supervision framework can effectively enhance various vision encoders' semantic understanding capabilities without requiring architectural changes or additional training objectives.
\section{Related Work}\label{related}
\paragraph{Vision-Language Pre-training} 
The integration of vision and language has led to Vision-Language Models (VLMs), which leverage image-text pairs to enrich semantic understanding. Contrastive learning has played a pivotal role in pre-training, with models like CLIP~\cite{CLIP}, ALIGN~\cite{ALIGN}, and SPARC~\cite{bica2024improving} applying softmax contrastive learning on large-scale datasets. Unlike these methods, SigLIP~\cite{SigLIP} introduces a simpler pairwise Sigmoid loss, removing the need for global similarity normalization. These models demonstrate strong zero-shot transfer capabilities, improving performance across multimodal tasks.

\paragraph{Cross-modal Alignment in MLLMs}
Cross-modal alignment in MLLMs typically follows deep or shallow fusion strategies. Deep fusion~\cite{flamingo, OBELICS, awadalla2023openflamingo, wang2023cogvlm} integrates vision encoder outputs into the LLM via interaction modules, allowing direct attention to image features. In contrast, shallow fusion~\cite{liu2024llava, FROMAGe, PaLM-E, blip2, minigpt4, bai2025qwen2,liu2023llava1.5} concatenates visual and text embeddings before passing them to the LLM, but struggles to bridge the alignment gap. Recent methods address this misalignment through techniques like similarity-based token assignment (AlignGPT~\cite{zhao2024aligngpt}) and segmentation/OCR-enhanced visual tokens (Rethinking MLLMs~\cite{lin2024rethinking}). However, these approaches fail to fundamentally improve adapter alignment. To address this, we propose Supervised Embedding Alignment (SEA), a token-level alignment paradigm that optimizes adapter integration for precise visual-text representation.
\section{Conclusion}
In this paper, we introduced Supervised Embedding Alignment (SEA), a token-level supervision alignment method that effectively bridges the modality gap in Multimodal Large Language Models.
By leveraging well-aligned vision-language models like CLIP, SEA provides precise semantic supervision for visual tokens, enabling their optimal alignment with the LLM's input space. Unlike conventional image-level alignment approaches, SEA mitigate both semantic distortion and modality representation gaps, substantially reducing the adaptation burden on language models during instruction-tuning.
SEA requires no additional data or inference cost, yet delivers consistent performance improvements across multiple benchmarks, with especially strong gains for smaller models. Our findings highlight the importance of token-level alignment for efficient multimodal learning and demonstrate that precise adapter design impacts both visual perception and language capabilities in MLLMs.

\section*{Limitations}
While SEA demonstrates strong performance in visual-textual integration, future directions could explore dynamic label selection that adapts to visual content complexity. Beyond images, extending this token-level alignment framework to other modalities (e.g., video, audio) while maintaining language model capabilities presents an important direction for developing general-purpose multimodal systems. The potential impact of new modality alignment paradigms on safety alignment remains an open question.

\bibliography{custom}

\newpage
\appendix
\appendix
\onecolumn
\renewcommand\thesubsection{\Alph{subsection}} 
\renewcommand\thefigure{\Alph{subsection}.\arabic{figure}} 
\renewcommand\thetable{\Alph{subsection}.\arabic{table}} 
\setcounter{figure}{0} 
\setcounter{table}{0}
\section*{Appendix}

\subsection{Experimental Setup}\label{appendix:exp}

\paragraph{Training details.}
We perform a two-stage training process. In the first stage, only the adapter was optimized while the vision encoder remained fixed. In the second stage, both the LLM and adapter were optimized. For SEA-PRIME, the vision encoder was also tuned in the second stage with a 2e-6 learning rate. We optimized all models for 1 epoch using the AdamW optimizer and a cosine learning schedule, following LLaVA's hyperparameters. The training time in ~\Cref{tab:ablation} ranges from 6 to 10 hours using 8$\times$H800 GPUs, nearly identical to LLaVA's training duration, with Stage 1 requiring only an additional 10-20 minutes. For SEA-PRIME, training takes less than 4 days with the same GPU configuration.

\paragraph{Datasets.}
For our models in ~\Cref{tab:main_result}, we use the Cambrian-1~\cite{tong2024cambrian1fullyopenvisioncentric} training data, which consists of 2.5M caption pairs for modality alignment and Cambrian-7M data for instruction tuning.
All ablation experiments in ~\Cref{tab:ablation} utilize the same data as LLaVA-1.5, specifically the CC-595K dataset~\cite{liu2024llava} for pre-training and a 656K mixture dataset~\cite{liu2023llava1.5}, which includes LLaVA-Instruct~\cite{liu2024llava}, TextVQA~\cite{textvqa}, GQA~\citep{hudson2019gqa}, OCR-VQA~\cite{mishra2019ocrvqa}, and Visual Genome~\cite{krishna2017vg} for instruction-tuning.

\subsection{Word List}\label{appendix:word_list}
We first performed syntactic analysis over the entire pretraining corpus to extract all meaningful and attribute-related words from the text. To expand coverage, we further incorporated frequent words from the 2of12 word list based on the Corpus 12 dictionary, resulting in a final vocabulary of approximately 4 million words. The LLaVA-Pretrain dataset was then processed using the pipeline illustrated in~\Cref{fig:method&pipeline}, where relevant semantic labels were assigned to each visual patch. As detailed in~\Cref{sec:method}, once the candidate semantic labels were defined, the similarity scores of all other words in the vocabulary were set to zero.


\subsection{Evaluating Alignment Consistency in Pretraining}\label{appendix:alignment_score}
During the pre-training, for a given image-text pair $(X_{\text{image}},X_{\text{text}})$. The LLM input is constructed as:
\begin{equation}
    X_{v} = g_{\theta}(f(X_{\text{image}})) \in R^{m\times dim}
\end{equation}
\begin{equation}
    X_{t} = \Psi(X_{\text{text}}) \in R^{n\times dim}
\end{equation}
where $f$ represents for vision encoder, $g$ represents for the adapter, and $\Psi$ is LLM's embedding layer. 
To quantify the alignment between visual and textual representations after the adapter, we introduce the \textbf{Token Alignment Consistency Score (TACS)}.
TACS is computed by measuring the cosine similarity between each visual token in the matrix \( X_v \) and each token in \( X_t \).
For each visual token, we identify the most similar text token based on similarity scores and record the similarity value. The final TACS score is obtained by averaging the top 10 highest similarity scores, providing a robust measure of alignment quality:
\begin{equation}
    \text{TACS} = \frac{1}{10} \sum_{i \in \text{Top 10}} \max_j \left( \frac{X_{v,i} \cdot X_{t,j}}{\|X_{v,i}\| \|X_{t,j}\|} \right)
\end{equation}

To create an evaluation dataset for assessing adapter alignment, we randomly selected 200 images from the COCO test dataset and manually annotated detailed captions for each image. As pretraining progresses, SEA achieves higher TACS scores, indicating improved alignment, while also showing corresponding improvements in POPE benchmark performance, as illustrated in Figure~\cref{fig:Aligment-POPE}.

\end{document}